%% file: main.tex
\documentclass[runningheads]{llncs}

\usepackage[mobile]{eccv}

\usepackage{eccvabbrv}

\usepackage{graphicx}
\usepackage{booktabs}

\usepackage[accsupp]{axessibility}  % Improves PDF readability for those with disabilities.

\usepackage[pagebackref,breaklinks,colorlinks,citecolor=eccvblue,linkcolor=eccvblue,urlcolor=eccvblue]{hyperref}
\usepackage{orcidlink}
\usepackage{epsfig}
\usepackage{graphicx}
\usepackage{amsmath}
\usepackage{amssymb}
\usepackage{adjustbox}
\usepackage[linesnumbered,boxed,ruled,commentsnumbered]{algorithm2e}
\usepackage{bbding}
\usepackage{pifont}
\usepackage{bm}
\usepackage{multirow}
\usepackage{multicol}
\usepackage{floatrow}
\usepackage[normalem]{ulem}
\usepackage{graphics}
\usepackage{algpseudocode}
\usepackage{caption}
\usepackage{makecell}
\usepackage{booktabs}
\usepackage{scalerel}
\usepackage{paralist}
\usepackage{color,colortbl}
\usepackage{amssymb}% http://ctan.org/pkg/amssymb
\usepackage{pifont}% http://ctan.org/pkg/pifont
\usepackage{markdown}
\newcommand{\cmark}{\ding{51}}%
\newcommand{\xmark}{\ding{55}}%

\definecolor{gray}{rgb}{0.5,0.5,0.5} 
\definecolor{frenchblue}{rgb}{0.0, 0.45, 0.73}
\definecolor{gray}{rgb}{0.5,0.5,0.5} 
\definecolor{green}{rgb}{0, 0.4, 0} 
\definecolor{orange}{rgb}{1, 0.5, 0} 	
\definecolor{mahogany}{rgb}{0.75, 0.25, 0.0}
\definecolor{purple}{rgb}{0.6, 0, 0.6}
\definecolor{darkgreen}{rgb}{0, 0.4, 0.4} 
\definecolor{teal}{rgb}{0.0, 0.5, 0.5}
\definecolor{aaaa}{rgb}{0.55, 0.1, 0.7}
\definecolor{red}{rgb}{1.0, 0, 0}
\definecolor{plotpurple}{rgb}{0.2353, 0.2, 0.90196}
\definecolor{plotorange}{rgb}{1.0, 0.6, 0.2}
\definecolor{plotgreen}{rgb}{0.2, 0.784313, 0.2}
\definecolor{plotred}{rgb}{1.0, 0.2, 0.392}
\definecolor{lightgray}{gray}{0.9}
\definecolor{LightCyan}{rgb}{0.88,1,1}
\definecolor{baselinecolor}{gray}{.9}
\definecolor{plotmagenta}{rgb}{0.839, 0, 1}

\newboolean{revising}
\setboolean{revising}{true}

\newcommand{\ap}{AP}

\newcommand{\apthreeD}{\ap$_{3\text{D}}$}

\newcommand{\apForty}{\ap$_{40}$}
\newcommand{\apEleven}{\ap$_{11}$}

\begin{document}

% ---------------------------------------------------------------
% TODO REVIEW: Replace with your title
\title{Weakly Supervised 3D Object Detection via Multi-Level Visual Guidance} 

% TODO REVIEW: If the paper title is too long for the running head, you can set
% an abbreviated paper title here. If not, comment out.
\titlerunning{VG-W3D}

% TODO FINAL: Replace with your author list. 
% Include the authors' OCRID for the camera-ready version, if at all possible.
\author{Kuan-Chih Huang\inst{1} \and
Yi-Hsuan Tsai\inst{2} \and
Ming-Hsuan Yang\inst{1,2}}

% TODO FINAL: Replace with an abbreviated list of authors.
\authorrunning{Huang et al.}
% First names are abbreviated in the running head.
% If there are more than two authors, 'et al.' is used.

% TODO FINAL: Replace with your institution list.
\institute{$^1$University of California, Merced \quad $^2$Google}

\maketitle

\begin{abstract}
  Weakly supervised 3D object detection aims to learn a 3D detector with lower annotation cost, e.g., 2D labels. 
    Unlike prior work which still relies on few accurate 3D annotations, we propose a framework to study how to leverage constraints between 2D and 3D domains without requiring any 3D labels.
    Specifically, we employ visual data from three perspectives to establish connections between 2D and 3D domains. 
    First, we design a feature-level constraint to align LiDAR and image features based on object-aware regions.
    Second, the output-level constraint is developed to enforce the overlap between 2D and projected 3D box estimations. 
    Finally, the training-level constraint is utilized by producing accurate and consistent 3D pseudo-labels that align with the visual data.
    We conduct extensive experiments on the KITTI dataset to validate the effectiveness of the proposed three constraints.
    Without using any 3D labels, our method achieves favorable performance against state-of-the-art approaches and is competitive with the method that uses 500-frame 3D annotations.
    %
    %Code and models will be made publicly available.
    Code will be made publicly available at: \url{https://github.com/kuanchihhuang/VG-W3D}.
\end{abstract}

\input{sec/0_Introduction}
\input{sec/1_RelatedWork}

\input{sec/2_Approach}

\input{sec/3_Experiments}
\input{sec/4_Conclusion}

\noindent{\textbf{Acknowledgements.}}
This work was supported in part by the Intelligence Advanced Research Projects Activity (IARPA) via Department of Interior/ Interior Business Center (DOI/IBC) contract number 140D0423C0074. The U.S. Government is authorized to reproduce and distribute reprints for Governmental purposes notwithstanding any copyright annotation thereon. Disclaimer: The views and conclusions contained herein are those of the authors and should not be interpreted as necessarily representing the official policies or endorsements, either expressed or implied, of IARPA, DOI/IBC, or the U.S. Government.

\bibliographystyle{splncs04}
\bibliography{main}
\end{document}

%% file: sec/0_Introduction.tex
\section{Introduction}

A key feature of autonomous systems is the ability to perceive and localize 3D objects accurately in the surrounding environments ~\cite{huang2023momam3t, huang2022monodtr, huang2021msanet, shi2019pointrcnn, huang2024ptt, chen2023voxenext}, enabling the agent (e.g., vehicles) to make informed decisions and navigate safely.
However, acquiring annotated 3D labels for training 3D object detection models poses challenges due to its cost and complexity, particularly when compared with the speed of labeling 2D boxes on visual data (e.g., 3-16 times slower, as shown in \cite{wei2021fgr}).
Hence, weakly supervised learning for 3D object detection has emerged as a practical approach to address the annotation bottleneck.

\begin{figure}
    \centering
    \includegraphics[width=0.6\linewidth]{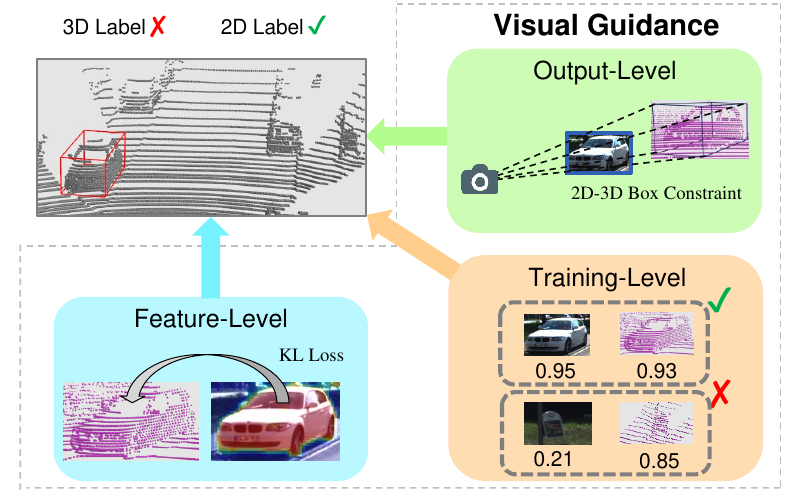}
    \caption{\textbf{Multi-level visual guidance for weakly-supervised 3D object detection.} We propose a framework to learn a 3D object detector from weak labels, e.g., 2D bounding boxes on the image plane, using three different perspectives, including feature-, output-, and training-level constraints. 
    Feature-level provides object-aware signals for point feature learning. 
    Output-level incorporates 2D-3D box constraints to enforce the model to generate reasonable box prediction. 
    Training-level guidance incorporates the confidence of 2D boxes into the pseudo-label technique to ensure the score consistency between 2D and 3D domains.
    }
    \label{fig:example}
    \vspace{-13pt}
\end{figure}

Recently, numerous approaches have been proposed to train 3D object detectors with reduced annotation requirements.
FGR \cite{wei2021fgr} proposes to generate 3D box candidates from the corresponding 2D boxes with frustum geometric relationships in a non-learning manner. 
WS3D \cite{meng2020ws3d, meng2021ws3d_pami} annotates the center of objects in BEV as weak labels and uses a few 3D labels for model training with cylindrical constraints. 
Several approaches \cite{MTrans, MAPGen} incorporate image and LiDAR information to learn the detectors jointly.
Nevertheless, existing methods either still require precise 3D box labels \cite{MTrans, MAPGen, meng2020ws3d, meng2021ws3d_pami}, or they only achieve suboptimal performance \cite{wei2021fgr} due to limited learning supervisions. 
Notably, none of these approaches explore the potential of incorporating various visual guidance across 2D and 3D during training.

In this work, we explore the integration of visual data into the training process of 3D object detectors, utilizing solely 2D annotations for weakly supervised 3D object detection, which is unique compared with the above-mentioned methods (see Table \ref{tab:comparison} for comparisons).
As shown in Figure \ref{fig:example}, we investigate the learning process with visual guidance from three perspectives: objectness learning from the feature level, response learning from the output level, and pseudo-label learning from the training level, with the following three observations.

{\it Observation 1: Feature-level Guidance.}
For the well-calibrated image and LiDAR, the objectness predictions obtained from the image should align with the corresponding regions in the LiDAR data.
For instance, when a point cloud is recognized as a foreground object by the 3D detector, its corresponding projected pixel on the image plane should align with a similar prediction made by the 2D detector, and vice versa.
Hence, we utilize this idea on the feature level by mapping image features to point features and considering only the objectness part to enhance the feature learning process of 3D object detectors.
% Utilizing a pre-trained image 2D detector from the dense visual domain allows us to leverage it to enhance the feature learning process of 3D object detectors.

{\it Observation 2: Output-level Guidance.} 
We notice a substantial overlap between the 2D and projected 3D bounding boxes on the image plane.
% (as shown in Figure \ref{fig:output}).
%
Based on this insight, we establish a distinct 2D-3D constraint to guide the supervision of 3D proposals. With this constraint, the model ensures that the estimated 3D box is accurately positioned within the frustum of the object's image region to generate higher-quality proposals.

{\it Observation 3: Training-level Guidance.} 
We find that using the initial 3D labels through a non-learning heuristic~\cite{wei2021fgr} can be noisy and partially missing objects due to the sparsity of point cloud data.
Hence, it is crucial to refine these labels iteratively for higher accuracy. 
Another challenge is to reduce false positives of generated pseudo-labels since during the self-training process, the model may easily produce unexpected estimations with high confidence scores.
% Utilizing the model's predictive 3D box results, pseudo-labels can be extended to unrecognized objects using a self-training approach. Despite this effort, generating high-quality labels remains challenging due to false positives with high confidence scores. 
%Therefore, we propose a solution: integrating prediction scores from the visual domain to refine 3D bounding boxes due to the better learning from image data with known 2D annotations.
%
Thus, we propose a solution by integrating prediction scores of 2D boxes from the visual domain into the pseudo-label technique to ensure score consistency for any object within both 2D and 3D domains. 

Based on these observations, we propose a 
multi-level visual guidance approach for weakly supervised 3D object detection, named VG-W3D.
We leverage visual cues to train a robust 3D object detector using only 2D annotations from three perspectives: feature, output, and training levels. 
Our method can further integrate with 2D annotations obtained from the off-the-shelf image object detectors trained from a different domain, making our method applicable to more scalable and cost-efficient 3D object detection.

Comprehensive experimental results validate the effectiveness of the proposed approaches. Specifically, when compared with methods in a similar annotation cost, our approach demonstrates a substantial improvement of at least 5.8\% in \apthreeD~on the KITTI~\cite{Geiger2012kitti} dataset.
Moreover, our method showcases comparable performance with state-of-the-art weakly supervised 3D detection methods that necessitate 500-frame 3D annotations.

\begin{table}[t]
\centering
\scriptsize
%\vspace{0.1cm} 
\setlength{\tabcolsep}{5pt}
	\begin{tabular}{lccccc}
	\toprule
	  \makecell[l]{Category} &\makecell[c]{Required\\3D Label} &\makecell[c]{Weak\\Label} &\makecell[c]{Guided\\by Output} &\makecell[c]{Guided\\by Feat.} &\makecell[c]{Guided\\by Training} \\
   \cmidrule(lr){1-1} \cmidrule(lr){2-6}
        VS3D \cite{qin2020vs3d} & \xmark  & Cls & \xmark  & \xmark & \xmark \\
        FGR \cite{wei2021fgr} &  \xmark  &2D Box & \cmark & \xmark & \xmark\\
        WS3D\cite{meng2020ws3d} & \cmark & BEV  & \xmark  & \xmark & \xmark \\
        MTrans \cite{MTrans} & \cmark & 2D Box & \cmark  & \xmark & \xmark \\ \midrule
        VG-W3D (Ours) & \xmark & 2D Box & \cmark  & \cmark & \cmark\\        
	\bottomrule
	\end{tabular}

\vspace{-0.25cm}
\caption{\textbf{Comparisons of VG-W3D and related work.} Our approach is guided by multiple visual cues, including 2D box outputs, image features, and scores for training, without requiring 3D annotation.}
%\vspace{-0.1cm}
 \label{tab:comparison}
\end{table}

%% file: sec/1_RelatedWork.tex
\section{Related Work}

\noindent {\bf LiDAR-based 3D Object Detection.}
LiDAR-based 3D object detectors \cite{shi2019pointrcnn, Lang2019pointpillars, he2020sassd} have gained much attention recently. 
Depending on different point aggregation techniques, they can be roughly classified into voxel and point-based methods. 
Voxel-based methods \cite{voxelnet, zheng2020ciassd, deng2020voxel, mao2021voxel} voxelize the point cloud as voxel grid representations, followed by the 2D and 3D convolution operations for object detection. 
In~\cite{voxelnet}, VoxelNet encodes the voxel features from the raw point cloud, followed by the dense region proposal network for 3D object detection.
CIA-SSD~\cite{zheng2020ciassd} utilizes a lightweight network on the bird's eye view (BEV) grid to learn spatial and semantic features and incorporates an IoU-aware confidence refinement module for robust detection.
Furthermore, Voxel-RCNN~\cite{mao2021voxel} applies voxel Region of Interest (RoI) pooling to extract voxel features within proposals for subsequent refinement,
while VoxelNeXt~\cite{chen2023voxenext} introduces a fully sparse architecture that directly predicts objects upon sparse voxel features instead of dense representations.

Numerous methods~\cite{shi2019pointrcnn, luo2021m3dssd, Point-GNN, PG-RCNN} directly utilize the raw point clouds as inputs to extract point-level features. 
PointRCNN \cite{shi2019pointrcnn} proposes a two-stage framework that segments the foreground points, generates the object-wise proposal, and then refines the canonical coordinates. 
On the other hand, 3DSSD~\cite{luo2021m3dssd} introduces a fusion sampling technique using feature distance to ensure comprehensive information preservation.
In~\cite{Point-GNN}, Point-GNN employs a graph neural network to create a more condensed representation of the point cloud data. 
Recently, Pointformer~\cite{Pan_2021_CVPR} uses a transformer module for local and global attention that operates directly on point clouds.

While these existing 3D object detectors have demonstrated desirable performance, they rely on 3D annotations for supervised learning, which is less efficient. In contrast, our approach emphasizes the development of a robust 3D detector without incurring labor-intensive costs.

\smallskip\noindent {\bf 3D Object Detection using Weak Labels.}
Annotating 3D bounding boxes on point clouds is time-consuming. 
As such, several methods focus on how to train a 3D detector with lower annotation costs to reduce the labor-intensive effort \cite{meng2020ws3d, meng2021ws3d_pami, wei2021fgr, MAPGen, MTrans, qin2020vs3d}. 
VS3D [49] produces 3D proposals by analyzing point cloud density in an unsupervised manner. Then, points within boxes are projected onto the image plane, guided by a 2D model trained with class labels to determine whether the proposal contains an object.
% VS3D [49] uses a 2D pre-trained model trained with the class-level label to supervise the 3D proposals, which are generated based on point cloud density, by recognizing whether the projected 2D box contains objects or not to finalize the final 3D boxes. 
%VS3D~\cite{qin2020vs3d} produces 3D proposals by analyzing point cloud density in an unsupervised manner. Subsequently, it projects these points contained within the boxes onto the image plane and utilizes a pretrained 2D model for object classification in the 2D plane, ultimately determining the final 3D boxes.
%
WS3D \cite{meng2020ws3d, meng2021ws3d_pami} utilizes center clicks to annotate the coarse location of the objects in BEV for producing the initial cylindrical object proposals, followed by the refinement process at the second stage with few accurate 3D labels.
In~\cite{wei2021fgr}, FGR introduces a non-learning technique that utilizes 2D bounding boxes to identify frustum sub-point clouds and then calculates the most precise 3D bounding box based on the segmented point cloud using heuristic methods.
MAP-Gen~\cite{MAPGen} and MTrans~\cite{MTrans} utilize rich image data to address the sparsity challenge inherent in 3D point clouds. However, most methods mentioned above still require partial centroid or accurate 3D annotation to develop their approaches.
In this work, we propose an approach that does not require any 3D annotations while still achieving competitive performance against existing weakly-supervised methods.

%% file: sec/2_Approach.tex
\section{Proposed Approach}
\subsection{Framework Overview}
Given a LiDAR point cloud $\mathcal{P}$ and its corresponding camera image $\mathcal{I}$, our goal is to develop a 3D object detector without using any 3D object annotation. 
In this paper, we introduce VG-W3D, a multi-level visual guidance approach for weakly-supervised 3D object detection, aiming to learn precise 3D bounding boxes using dense visual signals and annotated 2D boxes from the image domain.

\begin{figure*}[t]
\centering
\includegraphics[width=1\textwidth]{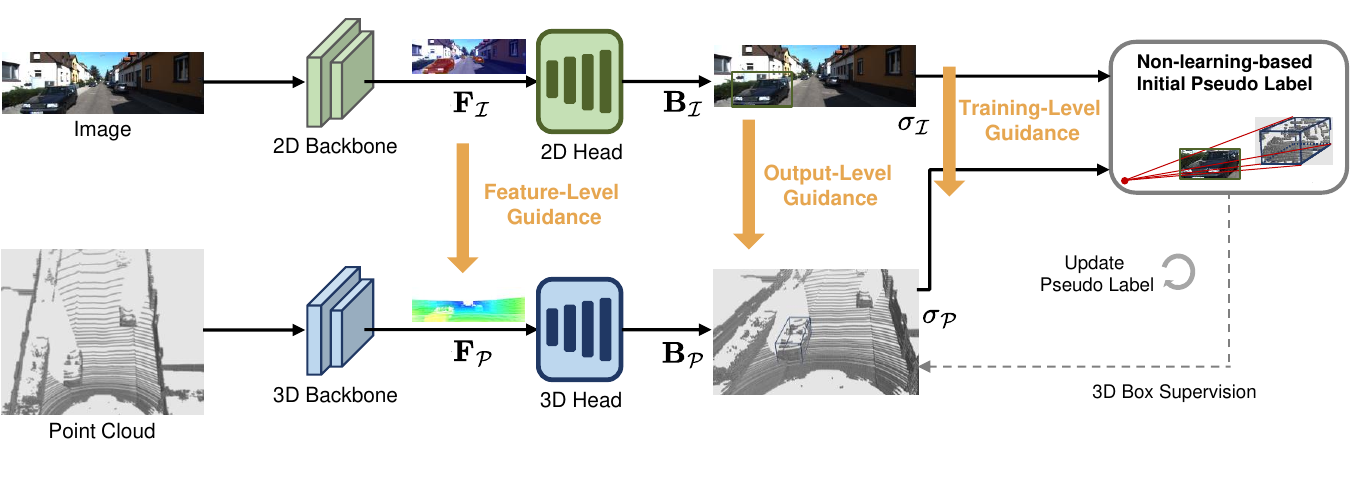}%with bbox line
\caption{\textbf{Overall framework of the proposed VG-W3D.} 
We utilize a non-learning method \cite{wei2021fgr} to identify frustum point clouds of objects, followed by a heuristic algorithm to estimate the initial noisy bounding boxes (top right).
In the image branch, we train an object detector based on 2D annotations to predict image features $\mathbf{F}_{\mathcal{I}}$ and 2D bounding boxes $\mathbf{B}_{\mathcal{I}}$ along with their confidence scores $\sigma_{\mathcal{I}}$, which serve as visual guidance for training the 3D detector.
Then, a PointNet-based 3D object detector is employed to extract point features $\mathbf{F}_{\mathcal{P}}$ and output 3D bounding boxes $\mathbf{B}_{\mathcal{P}}$ along with confidence scores  $\sigma_{\mathcal{P}}$. 
Our approach incorporates three levels of visual guidance for 3D training, namely feature-level (Section \ref{sec:feature}), output-level (Section \ref{sec:output}), and training-level (Section \ref{sec:train}). Note that the image branch is frozen during the training stage and is discarded during the inference stage.
}
\label{fig:arch}
\vspace{-20pt}
\end{figure*}

As illustrated in Figure \ref{fig:arch}, VG-W3D utilizes feature-level, output-level, and training-level cues to guide 3D object detection. 
To obtain initial 3D boxes for training, we adopt a non-learning approach similar to FGR \cite{wei2021fgr} to identify frustum point clouds of the objects, followed by a heuristic algorithm to estimate provisional 3D labels (see \cite{wei2021fgr} for more details).

In the initial phase, we train a 2D detector~\cite{zhou2019objects} using the provided 2D box annotations $\hat{\mathbf{B}}_\mathcal{I}$, to extract visual features $\mathbf{F}_{\mathcal{I}}$ and predict 2D bounding boxes $\mathbf{B}_\mathcal{I}$ along with their corresponding confidence scores $\sigma_\mathcal{I}$.
Subsequently, we employ a 3D PointNet-based detector~\cite{shi2019pointrcnn} to extract point cloud features $\mathbf{F}_{\mathcal{P}}$ and generate 3D bounding boxes $\mathbf{B}_\mathcal{P}$ with their detection scores  $\sigma_\mathcal{P}$.
 
In the following, we introduce three 2D visual cues to guide the 3D learning.
First, the feature-level cues constrain the attention towards foreground objects in images and point clouds, ensuring consistent predictions between the two modalities (Section~\ref{sec:feature}). 
Second, the output-level guidance ensures the logical positioning of predicted 3D proposals by enforcing the substantial overlaps between projected 3D boxes and 2D boxes on the image plane (Section~\ref{sec:output}). 
Finally, to produce high-quality pseudo-labels for retraining the 3D object detector, we incorporate image boxes and scores into the self-training technique, ensuring the improvement of pseudo-labels (Section~\ref{sec:train}).

\subsection{Feature-Level Visual Guidance}
\label{sec:feature}
%
%Intuitively, there is no guidance for each point's weakly-supervised 3D object detection setting. Though initial 3D bounding boxes are generated with a non-learning approach, they are noisy and contain lots of missing information. Conversely, we have dense image features training with complete 2D bounding boxes, which can generate meaningful guidance. 
%However, the bounding box is still coarse and not tight enough for point-level features. Thus, we propose to generate a point-level segmentation mask with self-supervised features, \, e.g., DINO features, which can help the model to learn the objectness of features. 
For weakly-supervised 3D object detection, individual points usually lack clear guidance. 
Even when initial 3D labels are available \cite{wei2021fgr}, it is still challenging to learn good 3D representations since these 3D labels are sometimes incomplete and noisy. 
Hence, we utilize dense visual hints to guide feature learning with known pixel-point mapping. 

\begin{figure}[!t]
    \centering
    \includegraphics[width=0.9\linewidth]{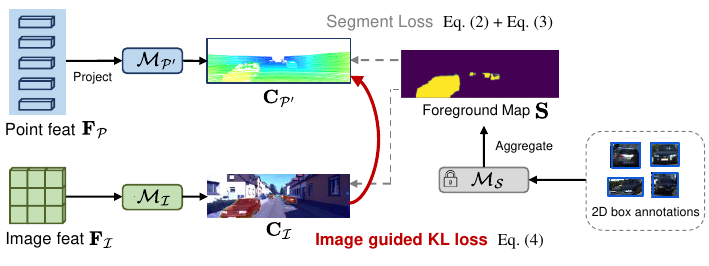}
    \caption{
    \textbf{
    Feature-level visual guidance.
    } Once we acquire the projected point features $\mathbf{F}_{\mathcal{P'}}$ and image feature $\mathbf{F}_{\mathcal{I}}$, we utilize object foreground map $\mathbf{S}$ to supervise the objections, in which the pretrained unsupervised instance segmentation module $\mathcal{M}_{\mathcal{S}}$ is applied to extract the object foreground map for each annotated 2D bounding box to generate $\mathbf{S}$.
    In addition, an image-guided KL divergence loss is applied to learn the distribution from the image features.
    }
    \label{fig:feature}
    \vspace{-4mm}
\end{figure}

%\vspace{1mm}
\smallskip \noindent {\bf Feature Mapping.}
As shown in Figure \ref{fig:feature}, we consider image features $\mathbf{F}_{\mathcal{I}} \in \mathbb{R}^{H \times W \times C}$ ($H \times W$ and $C$ denote the feature size and the number of feature channels) and point cloud features $\mathbf{F}_{\mathcal{P}} \in \mathbb{R}^{P \times C}$, with $P$ being the number of points. 
Initially, we project the point cloud features onto the image plane using the camera calibration parameters, yielding the projected point features: $\mathbf{F}_{\mathcal{P}'}=\rm{Proj}({\mathbf{F}_{\mathcal{P}}}) \in \mathbb{R}^{H \times W \times C}$. 
Consequently, we obtain the projected point cloud features matching pixel points on the image plane for better feature learning.

%\vspace{1mm}
\smallskip \noindent {\bf Feature Guidance.} 
One straightforward approach is enforcing point cloud features to mimic the image representations \cite{li2022uvtr} that learn the features of image modality from the LiDAR modality with L2 loss:
\begin{equation}
    \mathcal{L}_{feat}=\frac{1}{|\mathcal{A}|}\sum_{i \in \mathcal{A}}^{}
    {\lVert \mathbf{F}_{\mathcal{I}}(i) -     \mathbf{F}_{\mathcal{P}'}(i) \rVert}_2,
\end{equation}
where $\mathcal{A}$ is the valid pixel region on the image that matches with the projected points and ${\lVert \mathbf{\cdot} \rVert}_2$ is the $L_2$ norm distance. However, this may harm the procedure of point cloud feature learning since the image features cannot provide more geometric information than point clouds.
% based on the naive property of the difference between  these two modalities. 
Instead, we propose to enforce the predicted objectness probability for both image and point cloud features, $\mathbf{F}_{\mathcal{I}}$ and $\mathbf{F}_{\mathcal{P'}}$, to ensure that the 3D detector can recognize the foreground points with feature-level guidance.

%we observe that this can only achieve undesirable performance as this may limit the learning of the point cloud which have more geometric information (see Table \ref{tab:abl_feat}). 
%Instead, we propose to enforce the predicted objectness probability for both image and point cloud features, $\mathbf{F}_{\mathcal{I}}$ and $\mathbf{F}_{\mathcal{P'}}$ to ensure the 3D detector can recognize the foreground points. 
%

We employ the segmentation map for objectness supervision to allow guidance only in object regions. Without incurring additional annotation costs, we utilize self-supervised segmentation methods \cite{wang2023cut, niu2023unsupervised} to generate the foreground map without annotations. Specifically, within each ground truth 2D bounding box, objects are extracted along with their foreground maps for each object obtained by DINO~\cite{caron2021emerging}.
% and DINO~\cite{caron2021emerging} is applied to obtain the foreground region, creating a foreground map for each object.
These individual maps are then merged to form a segmentation ground truth map $\mathbf{S} \in \mathbb{R}^{H \times W}$.
% , where 1 represents foreground and 0 represents background.
%, which is used for learning the objectness of both point features 
%
We then exploit a classifier $\mathcal{M}_{\mathcal{P'}}$ to map the point cloud features to the binary probabilities that predict the objectness of the projected point cloud: $\mathbf{C}_{\mathcal{P'}}=\mathcal{M}_{\mathcal{P'}}(\mathbf{F}_{\mathcal{P'}})$ via the focal loss $\rm{FL}$:
\begin{equation}
\mathcal{L}_{seg}^{\mathcal{P}}=\frac{1}{|\mathcal{A}|}\sum_{i \in \mathcal{A}}^{}
    \rm{FL}(\mathbf{C}_{\mathcal{P'}}(i), \mathbf{S}(i)).
\label{eq:seg}
\end{equation}
% where $\mathcal{A}$ is the pixel region on the image, and
On the other hand, for the image domain, we add another branch on the 2D image detector with a linear classifier $\mathcal{M}_{\mathcal{I}}$ to learn the objectness for the pixels on the image: $\mathbf{C}_{\mathcal{I}}=\mathcal{M}_{\mathcal{I}}(\mathbf{F}_{\mathcal{I}})$, using a similar loss function to \eqref{eq:seg}:
\begin{equation}
    \mathcal{L}_{seg}^{\mathcal{I}}=\frac{1}{|\mathcal{A}|}\sum_{i \in \mathcal{A}}^{}
    \rm{FL}(\mathbf{C}_{\mathcal{I}}(i), \mathbf{S}(i)).
\label{eq:seg_i}
\end{equation}

%To generate more meaningful point cloud features, we propose to utilize image features to guide the learning of point features, which is different from some works [CITE] that learn the BEV features of image modality from LiDAR modality. 
%Besides, these works directly enforce the student model to mimic the feature maps from the teacher maps:
%\begin{align}
%    \mathcal{L}_{feat}=\frac{1}{|\mathcal{A}|}\sum_{i %\in \mathcal{A}}^{}
%    {\lVert \mathbf{F}_{\mathcal{I}}(i) -     %\mathbf{F}_{\mathcal{P}}(i) \rVert}_2
%\end{align}
%where ${\lVert \mathbf{\cdot} \rVert}_2$ is $L_2$ norm distance. However, this may damage the procedure of point cloud learning since the image features cannot provide more geometric features compared with Point cloud based on the naive property of the difference between these two modalities. 

To this end, we can utilize KL divergence loss to enforce the objectness in point cloud modality and learn the similar distribution as image modality without losing the geometric information from point clouds:
% , which contains complete data information:
\begin{equation}
    \mathcal{L}_{kl}= \rm{KL}(\mathbf{C}_{\mathcal{I}} || \mathbf{C}_{\mathcal{P'}}).
\label{eq:kl}
\end{equation}
%We utilize KL divergence loss to enforce the objectness in point cloud modality to learn the similar distribution as image modality containing more complete data information. See Table \ref{tab:feature} for more details.
% This can help the model learn representative point cloud features to achieve better performance.
% (see Table \ref{tab:abl_feat}). 

\subsection{Output-Level Visual Guidance}
\label{sec:output}
\begin{figure}[!t]
    \centering
\includegraphics[width=0.7\linewidth]{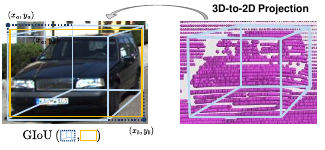}
    \caption{
    \textbf{
    Output-level visual guidance.
    } 
    The overlap of the projected 3D bounding box with the corresponding 2D bounding box signifies that 2D boxes can serve as supervision signals without 3D annotations. 
    We use GIoU loss to constrain the projected box of the learned 3D box on the image plane with the ground truth 2D object box.}
    \label{fig:output}
    \vspace{-4mm}
\end{figure}

It is worth mentioning that any object detected with both 2D and 3D bounding boxes should exhibit a high overlap~\cite{tang2019transferable3D,Park2022DetMatchTT}, as illustrated in Figure~\ref{fig:output}. This implies that in weakly-supervised learning scenarios where the 3D bounding box is unknown, we can utilize the ground truth 2D box to supervise the 3D boxes predicted by the 3D detector.

First, given the predicted 3D bounding box $\mathbf{B}_{\mathcal{P}}$, we obtain its eight corners in 3D coordinates denoted as $\mathcal{C}_3(\mathbf{B}_{\mathcal{P}}) \in \mathbb{R}^{8 \times 3}$. 
% Given the predicted 3D bounding box $\mathbf{B}_{\mathcal{P}}$ and its corresponding 2D bounding box $\mathbf{B}_{\mathcal{I}}$, the initial step involves transforming the 3D box to eight corners in 3D coordinates denoted as $\mathcal{C}_3(\mathbf{B}_{\mathcal{P}}) \in \mathbb{R}^{8 \times 3}$. 
%
Then, we obtain projected corners $\mathcal{C} \in \mathbb{R}^{8 \times 2}$ in 2D by using the known camera calibration parameters. 
Next, the bounding box of $\mathcal{C}$ can be determined by:
\begin{equation}
    (x_a,y_a) = {\rm min}(\mathcal{C}), (x_b,y_b) = {\rm max}(\mathcal{C}),
\end{equation}
% where $\rm{max}$ and $\rm{min}$ are the maximum and minimum operators; and 
where $(x_a,y_a)$ and $(x_b,y_b)$ are the top-left and bottom-right coordinates of the box $\mathbf{B}_{\rm{proj}}$.
As a result, we utilize the corresponding 2D box prediction $\mathbf{B}_{\mathcal{I}}$ to constraint the difference between these two boxes:
\begin{equation}
    \mathcal{L}_{box}= \hat{\sigma}_{\mathcal{I}}(1 - {\rm GIoU}(\mathbf{B}_{\mathcal{I}}, \mathbf{B}_{{\rm proj}})),
\label{eq:iou}
\end{equation}
where {\rm GIoU} is the generalized intersection over union \cite{giou} to guide box learning.
% and ensure that the projected 3D box overlaps optimally with the ground truth 2D box.
Compared with the normal IoU loss, GIoU loss can better alleviate the vanishing gradient issue of the non-overlapping case \cite{giou} between the projected 3D box and the ground truth 2D box.
%, particularly when the projected 3D box does not often overlap with the ground truth 2D box at all during the initial stage of model training.

%
In addition, $\hat{\sigma}_{\mathcal{I}} =  \sigma_{\mathcal{I}} / \sum_{i}^{N}{\sigma_{\mathcal{I}_i}}$ is the normalized score for each predicted 2D box across all $N$ objects in the same scene.
%
%Here, we consider the GIoU loss term rather than the conventional Intersection over Union (IoU) loss because the projected 3D box often does not perfectly overlap with the 2D box. Therefore, using GIoU would facilitate a more effective learning process.
% We observe that the GIoU loss outperforms the Intersection over Union (IoU) loss. This superiority is attributed to the fact that the projected 3D box often does not perfectly overlap with the 2D boxes, and GIoU facilitates a more effective learning process for such cases.
We observe that objects with low confidence by the 2D detector indicate the uncertainty of the predicted box, which may not be a high-quality box. Therefore, it is not suitable to treat each box equally. 
As such, we introduce the prediction score $\hat{\sigma}_{\mathcal{I}}$ as the weight of the loss in \eqref{eq:iou} for each box.
To this end, the proposed output-level guidance can ensure precise alignment between the projected bounding box and its 2D counterpart.

\IncMargin{1em}
\begin{algorithm}[!t]
    \footnotesize
    \SetAlgoNoLine 
    \SetKwInOut{Input}{\textbf{Input}}\SetKwInOut{Output}{\textbf{Output}} 
    \Input{
    Point cloud $\mathcal{P}$, initial 3D box $\hat{\mathbf{B}}_{0}$, pre-trained image detector $\Theta_\mathcal{I}$ with 2D box annotations $\hat{\mathbf{B}}_\mathcal{I}$, and predicted confidence $\sigma_{\mathcal{I}}$;\\
    %Calibrated Image $\mathcal{I}$ with 2D Box annotations $\hat{\mathbf{B}}_\mathcal{I}$;\\   
    }
    \Output{
        3D bounding box predictions $\mathbf{B}_\mathcal{P}$\\}
    \BlankLine
    %Initialize the coarse 3D box with Frustum segmentation and minimum bounding box estimation $\hat{\mathbf{B}}_{0}$; \\
    \Repeat
        {\text{convergence}}
        {
        Train a 3D detector $\Theta$ with $\hat{\mathbf{B}}_{t}$, $\hat{\mathbf{B}}_\mathcal{I}$ and $\Theta_\mathcal{I}$\\
        Obtain 3D pseudo-labels $\mathbf{B}_{t+1}$ with confidence $\sigma_{\mathcal{P}}$ \\
        %with image-point constraints from feature and output levels; \\
        Update pseudo-labels $\mathbf{B}_{t+1}$ to $\hat{\mathbf{B}}_{t+1}$ with: \newline
        1) $\mathbf{B}_{\rm{overlap}}=\{\mathbf{B}_{t+1}|\text{IoU}(\rm{Proj}(\mathbf{B}_{t+1}), \hat{\mathbf{B}}_\mathcal{I}) > \alpha_{0}, (\sigma_{\mathcal{P}} + \sigma_{\mathcal{I}})/2 > \alpha_{1}\}$ 
        \newline
        2) $\mathbf{B}_{\rm{unmatch}}= \mathbf{B}_{t+1}\symbol{92}\mathbf{B}_{\rm{overlap}}; \mathbf{B}_{\rm{score}} = \{\mathbf{B}_{\rm{unmatch}}|\sigma_{\mathcal{P}} > \alpha_{2}\}$
        \newline
        3)
        $\hat{\mathbf{B}}_{t+1} = \mathbf{B}_{\rm{overlap}} + \mathbf{B}_{\rm{score}}$\\
        $\mathbf{B}_\mathcal{P}$ = $\hat{\mathbf{B}}_{t+1}$
        }
    \caption{Training-level Guidance \label{selftrain}}
\end{algorithm}

%
%Therefore, we introduce the prediction score as the weight of the loss as in \eqref{eq:iou} for each box (weight will be normalized across all boxes in the same scene).

\subsection{Training-Level Visual Guidance}
\label{sec:train}
One common approach to providing direct supervisory signals is to use pseudo-labels. 
We initially consider other approaches like \cite{wei2021fgr} to produce 3D pseudo-labels without learning. 
However, they are prone to noise and may miss numerous objects.
% Despite having initial 3D labels generated through a non-learning approach \cite{wei2021fgr} for 3D training, they are still prone to noise and missing information.
%
For instance, only about 2700 frames contain pseudo-labels among the training dataset (3712 frames) generated by \cite{wei2021fgr}. 
%
% Thus, it is necessary to generate better pseudo-labels to retrain the model.
%
Moreover, pseudo-labels may introduce additional false positives that 
may negatively affect self-training.
% While pseudo-labeling can generate labels for retraining the model, it may introduce additional false positives, potentially resulting in even worse performance 
% (see Table~\ref{tab:abl_train}). 
%
To handle these issues, we introduce an image-guided approach to generate high-quality pseudo-labels, outlined in Algorithm \ref{selftrain}.
The method starts with initial 3D pseudo-labels $\hat{\mathbf{B}}_{0}$ from \cite{wei2021fgr}, 2D box annotation $\mathbf{B}_\mathcal{I}$, and a pre-trained 2D object detector $\Theta_\mathcal{I}$.
We utilize a 2D detector to predict the confidence score $\sigma_{\mathcal{I}}$ for each 2D annotation. In practice, we extract the confidence of the center index for each ground truth object from the heatmap predicted by the image detector. 

Each iteration consists of three main steps.
First, we train the 3D object detector based on pseudo-labels $\hat{\mathbf{B}}_{t}$ at round $t$ and use the feature- and output-levels guidance as mentioned in Section \ref{sec:feature} and \ref{sec:output}, respectively. 
Second, we generate the initial pseudo-labels $\hat{\mathbf{B}}_{t+1}$ along with corresponding confidence ${\sigma}_{\mathcal{P}}$ for the next round.
Finally, we filter pseudo-labels based on the following criteria to ensure the quality: 
1) Utilizing the Hungarian algorithm, we match 2D and projected 3D bounding boxes, retaining those with an IoU score greater than the threshold $\alpha_0$.
Moreover, the averaged confidence score from 2D and 3D boxes should be larger than $\alpha_1$. The generated pseudo-labels are denoted as $\mathbf{B}_{\rm{overlap}}$.
% and those confidence scores add image confidence should be larger than $2\alpha_1$, denoted as $\mathbf{B}_{\rm{overlap}}$. 
%
2) For the remaining projected 3D boxes $\mathbf{B}_{\rm{unmatch}}=\mathbf{B}_{t+1}\symbol{92}\mathbf{B}_{\rm{overlap}}$, we apply the non-maximum suppression (NMS) to eliminate redundant boxes, retaining only those with high confidence $\alpha_2$, termed as $\mathbf{B}_{\rm{score}}$.
The final selected 3D pseudo-labels are the two sets: $\hat{\mathbf{B}}_{t+1}=\mathbf{B}_{\rm{overlap}}+\mathbf{B}_{\rm{score}}$.

\subsection{Training Objectives}
\label{sec:loss}
We apply two loss functions to train our weakly-supervised 3D object detection framework. For scenes with the 3D pseudo-labels:
\begin{equation}
\mathcal{L}_{pl}=\mathcal{L}_{rpn}+\mathcal{L}_{rcnn}+\mathcal{L}_{seg}^{\mathcal{P}}+\mathcal{L}_{kl},
\label{eq:loss}
\end{equation}
where $\mathcal{L}_{rpn}$ and $\mathcal{L}_{rcnn}$ represent the loss functions of RPN and refinement network proposed in~\cite{shi2019pointrcnn}, respectively. $\mathcal{L}_{seg}^{\mathcal{P}}$ and $\mathcal{L}_{kl}$ are the feature-level guidance as in \eqref{eq:seg} and \eqref{eq:kl}.
For scenes with only 2D labels, the weakly-supervised loss functions are applied:
\begin{equation}
\mathcal{L}_{weak}=\mathcal{L}_{seg}^{\mathcal{P}}+\mathcal{L}_{kl}+\mathcal{L}_{box},
\label{eq:weak_loss}
\end{equation}
where $\mathcal{L}_{box}$ is the loss of box-level guidance defined in \eqref{eq:iou}.
For the 2D object detector, we adopt the same loss function as CenterNet~\cite{zhou2019objects} with an additional loss $\mathcal{L}_{seg}^{\mathcal{I}}$ in \eqref{eq:seg_i}.

%\subsection{Ne Details}

%% file: sec/3_Experiments.tex
\section{Experiments}
\subsection{Experimental Setup}
\label{sec:setup}
\noindent{\bf {Datasets.}}
We evaluate the proposed method on the KITTI 3D object detection~\cite{Geiger2012kitti} dataset,
% and nuScenes~\cite{nuscenes2020caesar} datasets.
which comprises 7481 images for training and 7518 images for testing. Following the protocol established in \cite{chen2015_3dop}, we divide the training samples into a training set (3712) and a validation set (3769) for conducting experiments for the validation set and ablation studies.
%More results can be found in the supplementary materials.
% are performed based on this division.
%For the nuScenes dataset, we leave the details and results in the {\color{blue} supplementary material}.

\input{tab/0_kitti_test}

% \vspace{1mm}
\smallskip\noindent{\bf Evaluation Metrics.}
We assess performance using the average precision (AP) metric for 3D object detection and bird's eye view (BEV) detection tasks. To mitigate bias \cite{andrea2019monodis}, we adopt the 40 recall positions metric (\apForty) instead of the original 11 (\apEleven). The benchmark categorizes detection difficulty into three levels ("Easy," "Mod.," "Hard"), considering size, occlusion, and truncation. All methods are ranked based on \apthreeD~ score in the moderate setting (Mod.), aligning with the KITTI benchmark.

% \vspace{1mm}
\smallskip\noindent{\bf Implementation Details.}
We adopt CenterNet~\cite{zhou2019objects} as the 2D detector trained for 140 epochs with an extra head for foreground map segmentation in \eqref{eq:seg_i}. On the other hand, we utilize PointRCNN~\cite{shi2019pointrcnn} as the 3D detector for fair comparisons, trained for 30 epochs. For output-level guidance, we choose the overlapped bounding box with IoU larger than 0.5 for training. For training-level guidance, we set $\alpha_0=0.5$, $\alpha_1=0.5$ and $\alpha_2=0.95$.
We will make the source codes and models available to the public.
% For data augmentation, we apply random cropping and scaling to train the image detector, and random flip, scaling, and rotating to train the 3D detector.

\subsection{Main Results}

% \vspace{-1mm}
\noindent{\bf KITTI test set.}
Table \ref{tab:kitti_test} compares our VG-W3D with several state-of-the-art weakly-supervised 3D object detection methods and fully-supervised baselines on the KITTI test set.
%
%Our method demonstrates superior performance compared to other approaches.
%
Compared with the fully-supervised baseline, PointRCNN, our VG-W3D achieves competitive performance without using any 3D annotations, demonstrating the effectiveness of our method. 
Furthermore, compared with our weakly-supervised baseline, FGR~\cite{wei2021fgr}, our approach obtains improvements by \textbf{3.83/5.81/6.33} in \apthreeD~at IoU threshold 0.7 on three settings, which show the effectiveness of the proposed visual guidance approach. Furthermore, compared with methods (e.g., WS3D~\cite{meng2020ws3d, meng2021ws3d_pami}) that require 3D or BEV central annotations, our method performs favorably against these approaches.

\input{tab/kitti_tracking_val}
\input{tab/abl_three}

% \vspace{1mm}
\smallskip\noindent{\bf KITTI val set.}
We also conduct experiments on the KITTI validation set in Table~\ref{tab:kitti_val}.
Similarly, our approach achieves favorable performance against several
weakly-supervised methods.
Specifically, compared to MTrans~\cite{MTrans} that requires 500-frame 3D annotations for training, our VG-W3D with the same 3D detector (i.e., PointRCNN~\cite{shi2019pointrcnn}) achieves better results on most metrics, which validates the effectiveness of our approach.
Furthermore, compared with VS3D~\cite{qin2020vs3d} and WS3DPR~\cite{lie2022WS3DPR} that only utilize class-level labels for classifying the proposal, our approach can effectively guide the learning from multiple perspectives, resulting in significant performance improvement.
\subsection{Ablation Study and Analysis}
\label{sec:ablation}

\noindent{\bf Importance of different levels of visual guidance.}
In Table \ref{tab:abl_diff}, we show the effectiveness of different levels of visual guidance. 
Without any proposed visual guidance, 
the baseline (a) we reproduce via FGR~\cite{wei2021fgr} only achieves undesirable performance since the initial 3D annotations from the non-learning method are noisy.
After adding the training-level constraint in row (b), the model is able to learn the objectness of the points.
% , which helps to improve the performance. 
%
Utilizing the proposed output-level guidance in row (c) can further enforce the model to generate reasonable proposals to enhance detection ability.
% (a$\rightarrow$ c). 
%
We also show the benefit of retraining the model with our training-level guidance in row (d).
% (a $\rightarrow$ d).
%
Finally, the results in rows (e) and (f) demonstrate the efficacy of combining visual guidance from feature-, output-, and training-levels.

%\vspace{1mm}
\smallskip{\noindent{\bf Designs in feature-level guidance.}}
We investigate the effectiveness of the proposed feature-level guidance in Table \ref{tab:abl_feat}\footnote{To better understand the effect of feature- and output-level guidance, we do not include the training-level guidance that involves multiple retraining iterations in the ablation studies.}.
In the first row, a performance drop is observed when the objectness probability learned from the image domain is omitted. 
In addition, substituting the KL divergence loss with L2 loss and attempting to mimic features instead of objectness prediction yields unsatisfactory results, aligning with our expectations outlined in Section~\ref{sec:feature}.
Finally, using the 2D bounding box as a mask for foreground supervision is not effective, as it fails to provide valuable signals for point clouds, e.g., certain regions inside the box might belong to the background.

\input{tab/abl_feature} 
\input{tab/abl_output}

%\vspace{1mm}
\smallskip{\noindent{\bf Designs in output-level guidance.}}
In Table \ref{tab:output}, various options for output-level guidance are compared.
%
%Initially, we use the image detector to predict 3D box corners as guidance signals for 3D learning. 
Instead of regressing the IoU between 2D and 3D bounding boxes, we can utilize the L1 loss to learn the corners of the 3D box predicted from the pre-trained image detector.
However, this approach is less effective due to the noisy nature of initial 3D box annotations, impacting the ability of the image detector to learn the corners as guidance.
Furthermore, our results indicate that the Generalized Intersection over Union (GIoU) loss yields better performance than the Intersection over Union (IoU) loss, since the projected 3D box often does not perfectly overlap with the 2D box.

%\vspace{1mm}
\smallskip\noindent{\bf Designs in training-level guidance.}
We explore the effectiveness of various training-level guidance designs in Table \ref{tab:abl_train}. Initially, we employ the pseudo-labeling (PL) technique to re-generate labels using the trained model. 
However, this approach yields less favorable results as the model generates numerous false positives, causing some negative effects in the training process. 
Adding the constraints of overlapping the projected 3D box and ground truth 2D box ($\mathbf{B}_{\rm{overlap}}$ in Algorithm~\ref{selftrain}) slightly helps self-training.
% performance by about 1 \apthreeD. 
%
Finally, integrating two of our training-level guidance (i.e., $\mathbf{B}_{\rm{overlap}} + \mathbf{B}_{\rm{score}}$ in Algorithm~\ref{selftrain}) preserves predictions with high confidence scores, enables us to identify more objects lacking annotations in the image domain. This significantly improves the pseudo-labeling process.

\input{tab/abl_training}
\input{tab/pseudo_label_plus_coco}

\begin{figure*}[t]
\centering
\includegraphics[width=0.88\textwidth]{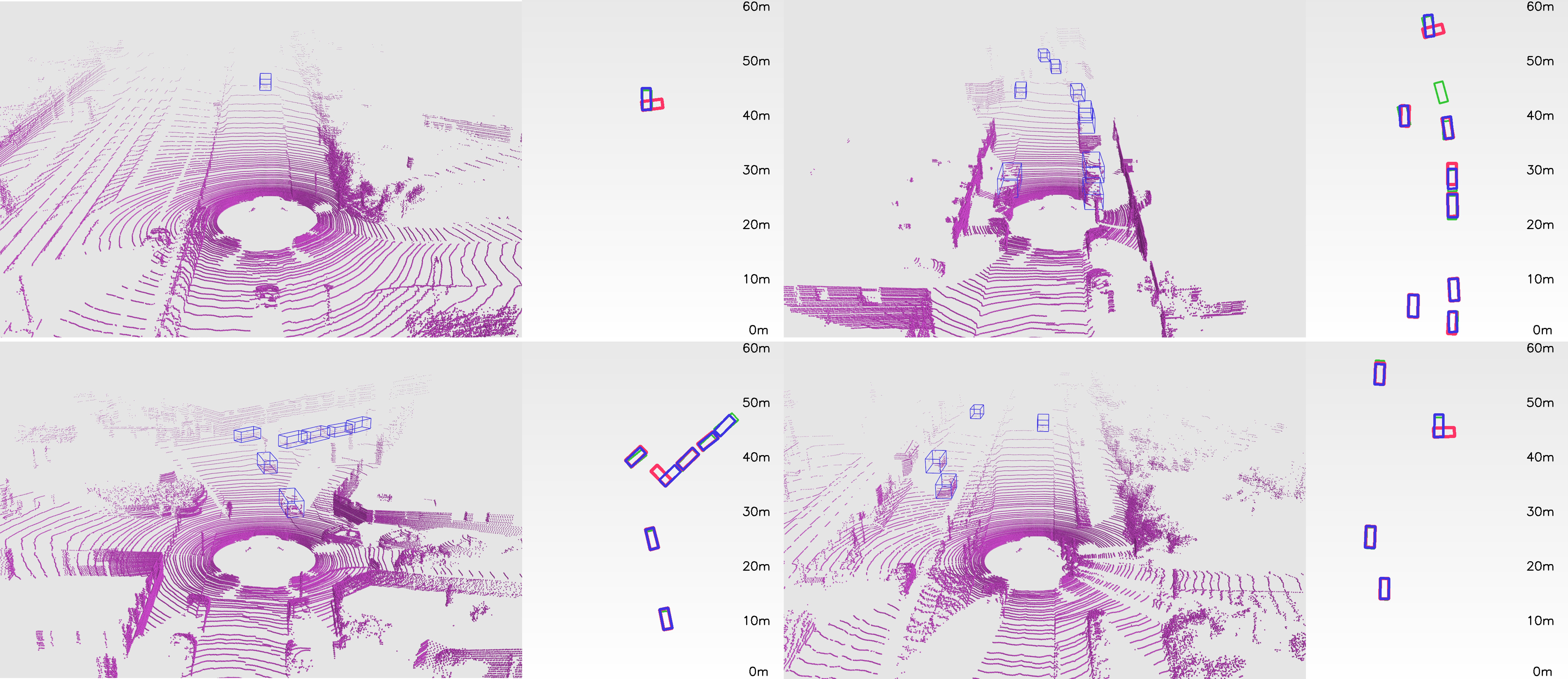}
% \vspace{-2mm}
\caption{\textbf{Qualitative visualizations on the KITTI validation set.} 
We provide the predictions on the LiDAR view (left) and bird's eye view (right) in each result. The \textcolor{plotpurple}{purple} boxes in the LiDAR data and BEV plane indicate the predictions from our VG-W3D. The \textcolor{plotgreen}{green} and \textcolor{plotred}{pink} boxes on BEV are the ground truth and predictions from the FGR~\cite{wei2021fgr} baseline (without the proposed visual guidance), respectively.
We show that our predictions have a closer alignment with the ground truth, while FGR often fails to enclose the point cloud tightly.
%
%Best viewed in color with zooming in.
}
\label{fig:kitti_vis}
\vspace{-2.5mm}
\end{figure*}

\smallskip\noindent{\bf Quality of pseudo-labels.} In Table~\ref{tab:quality}, we investigate the quality of the initial pseudo-label~\cite{wei2021fgr} and our proposed pseudo-label approach. 
The 3D labels generated through the non-learning approach often contain noise and missing information, particularly when the frustum point cloud lacks clear separation to locate the bounding box accurately.
% and the hierarchical estimation methods struggle to locate the bounding box accurately.
Therefore, the recall only achieves 46.71\% at IoU=0.7.
After the first round of training with our approach, the quality of the pseudo-label is improved by approximately 25\% recall, showcasing the effectiveness of our approach. 
Finally, the recall improvement of the pseudo label saturates after the second-round training.

%\input{tab/abl_coco}

%\vspace{-0.5mm}
\smallskip\noindent{\bf Cooperate with pre-trained 2D object detector.}
Our approach is able to cooperate with existing pre-trained off-the-shelf 2D object detectors, as shown in Table~\ref{tab:coco}. 
With the DETR~\cite{Nicolas2020detr} object detector trained on the COCO dataset~\cite{coco}, we can directly use it to detect the objects on the KITTI dataset and generate 2D boxes. 
Initially, we generate 3D bounding boxes using \cite{wei2021fgr} and train our model, establishing baseline results. 
Subsequently, we apply our proposed visual guidance approach. 
It is worth noting that our method achieves competitive performance compared to using 2D annotations. 
This demonstrates the potential scalability and effectiveness of the proposed framework.

\subsection{Qualitative Results}
We show qualitative examples of the KITTI validation set in Figure~\ref{fig:kitti_vis}.
In the results, we show predictions of our method both on the LiDAR and BEV (denoted as purple boxes), while results of FGR~\cite{wei2021fgr} and ground truths on BEV are denoted as pink and green boxes, respectively.

Compared to FGR~\cite{wei2021fgr} which is only trained with initial 3D labels, the predictions from our VG-W3D show a closer alignment with the ground truth. This validates the efficacy of the three proposed visual guidances in enhancing object detection performance. 
In addition, we observe that the non-learning approach typically fails to enclose the point cloud tightly, resulting in labels with inaccurate heading and consequently producing incorrect guidance.

%% file: tab/0_kitti_test.tex
\begin{table*}[t]
    \centering
    \scriptsize
    \renewcommand{\arraystretch}{1.1}
    \setlength{\tabcolsep}{1.5pt}
    \begin{tabular}{lccccc@{\hspace{4pt}}ccc}
    \toprule
    \multirow{2}{*}{Method}& \multirow{2}{1.5cm}{\centering Weak Labels} & \multirow{2}{1.7cm}{\centering Required 3D Annotations} & \multicolumn{3}{c}{\apthreeD@IoU=0.7} & \multicolumn{3}{c}{\apBev@IoU=0.7} \\
    \cmidrule(lr){4-6} \cmidrule(lr){7-9} & & & Easy & Mod. & Hard & Easy & Mod. & Hard\\
    \midrule
    PointPillars \cite{Lang2019pointpillars} & - & 	\cmark & 82.58 & 74.31 & 68.99  &  90.07 & 86.56 & 82.81\\
    PointRCNN \cite{shi2019pointrcnn} & - & \cmark & 86.96 & 75.64 & 70.70 & 92.13 & 87.39 & 82.72 \\
    %PV-RCNN \cite{shi2020pv} & - & \cmark & 90.25 & 81.43 & 76.82 & 94.98 & 90.65 & 86.14 \\
    \midrule
    WS3D \cite{meng2020ws3d} & BEV Centroid & 534o & 80.15 & 69.64 & 63.71 &  90.11 &84.02 & 76.97\\
    WS3D (2021)~\cite{meng2021ws3d_pami} & BEV Centroid & 534o  & 80.99 & 70.59 & 64.23 &  90.96 & 84.93 & 77.96 \\
    MAP-Gen~\cite{MAPGen} & 2D Box & 500f  & 81.51 & 74.14 & 67.55 &  90.61 & 85.91 & 80.58 \\
    MTrans (PointPillars)~\cite{MTrans} & 2D Box & 500f & 77.65 & 67.48 & 62.38 & - & - & - \\
    %MTrans (PointPillars) & 2D Box & 125f & 83.70 & 71.66 & 66.67 & - & - & -\\
    MTrans (PointRCNN)~\cite{MTrans} & 2D Box & 500f & 83.42 & 75.07 & 68.26 &  91.42 & 85.96 &78.82\\
    %MTrans & 2D Box & 125f & 87.64 & 77.31 & 74.32 & - & - & -\\
    %CAT~\cite{cat} & 2D Box & 500f  & 84.84 & 75.22 &70.05 &  91.48 & 85.97 & 80.93\\
    \midrule
    %WS3DPR~\cite{lie2022WS3DPR} & Cls & \xmark & 46.57 & 33.92 & 26.97 & - & - & - \\
    FGR~\cite{wei2021fgr}   & 2D Box & \xmark & 80.26 & 68.47 & 61.57 &  90.64 & 82.67 &75.46 \\
    %Ours(no self) & 2D Box &  \xmark & 81.40 & 69.23 &	63.57 & 90.92 &	82.63 &	77.23 \\
    %Ours & 2D Box & \xmark & 83.30 & 73.75 & 66.95 & 91.46 & 84.95 & 79.84 \\
    %Ours & 2D Box & \xmark & \textbf{82.12} & \textbf{74.10} &	\textbf{67.61} & \textbf{92.12} & \textbf{85.89} & \textbf{80.77} \\
    VG-W3D (Ours) & 2D Box & \xmark & \textbf{84.09} & \textbf{74.28} &	\textbf{67.90} & \textbf{91.88} & \textbf{85.61} & \textbf{80.67} \\
    \bottomrule
    \end{tabular}
    \vspace{-2.5mm}
    \caption{\textbf{Detection performance of the Car category on the KITTI test set.} We compare our VG-W3D with fully-supervised baselines (first block), weakly-supervised methods that require accurate 3D annotations (second block), and weakly-supervised baselines without any 3D annotation (last block). 500f and 534o denote that the methods require annotating 500 frames and 534 objects (about 120 frames) for training, respectively.}
    \label{tab:kitti_test}
\end{table*}

%% file: tab/kitti_tracking_val.tex
\begin{table}[t]
    \centering
    \scriptsize
    \renewcommand{\arraystretch}{1.1}
    \setlength{\tabcolsep}{8pt}
    \begin{tabular}{lcccc}
    \toprule
    \multirow{2}{*}{Method}&  \multirow{2}{1.7cm}{\centering Required 3D Annotations} & \multicolumn{3}{c}{\apthreeD@IoU=0.7}  \\
    \cmidrule(lr){3-5}  & & Easy & Mod. & Hard \\
    \midrule
    PointPillars \cite{Lang2019pointpillars} & 	\cmark & 86.10 & 76.58 & 72.79 \\
    PointRCNN \cite{shi2019pointrcnn} & \cmark & 88.99 & 78.71 & 78.21  \\
    %PV-RCNN \cite{shi2020pv}  & \cmark & 90.25 & 81.43 & 76.82  \\
    \midrule
    WS3D \cite{meng2020ws3d} & 534o & 84.04 & 75.10 & 73.29 \\
    WS3D (2021)~\cite{meng2021ws3d_pami} &  534o  &  85.04 & 75.94 & 74.38  \\
    MAP-Gen~\cite{MAPGen} &  500f  & 87.87 & 77.98 & 76.18  \\
    MTrans~\cite{MTrans}  & 125f & 87.64 & 77.31 & 74.32 \\
    %MTrans (PointPillars) & 2D Box & 125f & 83.70 & 71.66 & 66.67 & - & - & -\\
    MTrans~\cite{MTrans}  & 500f & 88.72 & 78.84 & 77.43 \\
    %CAT~\cite{cat}  & 500f  & 89.19 &79.02 & 77.74 \\
    \midrule
    VS3D~\cite{qin2020vs3d}& \xmark & 9.09 & 5.73 & 5.03 \\
    WS3DPR~\cite{lie2022WS3DPR}& \xmark & 60.01 & 44.48 & 36.93 \\
    FGR~\cite{wei2021fgr}  & \xmark & 86.68 & 73.55 & 67.91 \\
    FGR~\cite{wei2021fgr}* & \xmark & 87.19 & 74.00 & 68.34 \\
    VG-W3D (Ours) &  \xmark & \textbf{91.32} & \textbf{78.89} & \textbf{74.70}\\ %[24 repro/19]
    \bottomrule
    \end{tabular}
    \vspace{-2.5mm}
    \caption{\textbf{Detection performance of the Car category on the KITTI validation set.} We compare our VG-W3D with fully-supervised and weakly-supervised methods. 
    * means the results are reproduced by ourselves.}
    \label{tab:kitti_val}
\end{table}

%% file: tab/abl_three.tex
\begin{table}[t]
    \centering
    \scriptsize
    \renewcommand{\arraystretch}{1.1}
    \setlength{\tabcolsep}{6pt}
    \begin{tabular}{lccccccc}
    \toprule
    \multirow{2}{*}{}& \multirow{2}{*}{Feature} & \multirow{2}{*}{Output} & \multirow{2}{*}{Training} &  \multicolumn{3}{c}{\apthreeD@IoU=0.7}  \\
    \cmidrule(lr){5-7} & & & & Easy & Mod. & Hard \\
    \midrule
    (a) & \xmark & \xmark & \xmark & 87.19 & 74.00 & 68.34  \\
    (b) & \cmark & \xmark & \xmark & 89.12 &74.29 & 70.78  \\ %[23]
    %(e) & \cmark & \cmark & \xmark & 88.79 & 76.15 & 71.49  \\
    (c) & \xmark & \cmark & \xmark & 88.95 & 76.42 & 71.58  \\
    (d) & \xmark & \xmark & \cmark & 88.95 & 77.75 & 73.31  \\ %[25]
    (e) & \cmark & \cmark & \xmark & 89.43 & 76.71 & 72.06 \\ %[22]
    (f) & \cmark & \cmark & \cmark & \textbf{91.32} & \textbf{78.89} & \textbf{74.70} \\
    \bottomrule
    \end{tabular}
    \vspace{-2.5mm}
    \caption{\textbf{Ablation study with different levels of visual guidance in our proposed framework} on the KITTI validation set.}
    \label{tab:abl_diff}
\end{table}

%% file: tab/abl_feature.tex
\begin{table}[t]
    %\vspace{-1mm}
    \centering
    \scriptsize
    \renewcommand{\arraystretch}{1.1}
    \setlength{\tabcolsep}{8pt}
    \begin{tabular}{llcccc}
    \toprule
      \multirow{2}{*}{Loss}& \multirow{2}{*}{Mask}& \multicolumn{3}{c}{\apthreeD@IoU=0.7}  \\
    \cmidrule(lr){3-5} & & Easy & Mod. & Hard \\
    \midrule
    \xmark & Segment & 88.60 & 74.61 & 71.10 \\
    L2 & Segment & 88.87 & 75.69 & 71.34 \\
    KL & 2D Box & 89.13 & 74.65 & 69.88 \\
    \midrule
    KL & Segment & 89.43 & 76.71 & 72.06  \\
    \bottomrule
    \end{tabular}
    \vspace{-2.5mm}
    \caption{\textbf{Ablation study with different designs of feature-level visual guidance} on the KITTI validation set.}
    \label{tab:abl_feat}
\end{table}

%% file: tab/abl_output.tex
\vspace{-0.5mm}
\begin{table}[t]
    % \vspace{2mm}
    \centering
    \scriptsize
    \renewcommand{\arraystretch}{1.1}
    \setlength{\tabcolsep}{4pt}
    \begin{tabular}{lccccccc}
    \toprule
      \multirow{2}{*}{2D-3D loss}& \multicolumn{3}{c}{\apthreeD@IoU=0.7} & \multicolumn{3}{c}{\apBev@IoU=0.7} \\
    \cmidrule(lr){2-4} \cmidrule(lr){5-7} & Easy & Mod. & Hard & Easy & Mod. & Hard\\
    \midrule
    L1 & 88.81 & 74.70 &71.36 &94.56 & 85.05 & 80.84 \\ %[5]
    IoU  & 88.51 & 74.50 & 71.40 & 95.16 & 85.71 & \textbf{82.15}\\
    GIoU & \textbf{89.43} & \textbf{76.71} & \textbf{72.06} & \textbf{95.51} & \textbf{86.31} & 81.84 \\
    \bottomrule
    \end{tabular}
    \vspace{-2.5mm}
    \caption{\textbf{Ablation study with different 2D-3D constraint loss of our output-level visual guidance} on the KITTI validation set. 
    %'Keypoints' means predicting the keypoints from the image plane and supervising the corresponding from the LiDAR outputs.
    }
    \label{tab:output}
\end{table}

%% file: tab/abl_training.tex
%\vspace{-0.5mm}
\begin{table}[t]
    \centering
    \scriptsize
    \renewcommand{\arraystretch}{1.1}
    \setlength{\tabcolsep}{4pt}
    \begin{tabular}{lccccccc}
    \toprule
      \multirow{2}{*}{Method}& \multicolumn{3}{c}{\apthreeD@IoU=0.7} & \multicolumn{3}{c}{\apBev@IoU=0.7} \\
    \cmidrule(lr){2-4} \cmidrule(lr){5-7} & Easy & Mod. & Hard & Easy & Mod. & Hard\\
    \midrule
    PL & 91.06 & 76.64 & 73.79 & 95.24 & 86.06 & 84.32   \\
    \midrule
    Overlap & 91.07 & 77.55 & 73.74 & 95.21 & 86.01 & 84.28   \\
    Overlap + Score & 91.31 & 78.89 & 74.70 & 95.50 & 88.16 & 85.99 \\ %[19]repro_v2
    \bottomrule
    \end{tabular}
    \vspace{-2.5mm}
    \caption{\textbf{Ablation study with different designs of training-level guidance} on the KITTI validation set. ``PL'' denotes the pseudo-labeling technique, while ``Overlap'' and ``Score'' are $\mathbf{B}_{\rm{overlap}}$ and $\mathbf{B}_{\rm{score}}$ in Algorithm~\ref{selftrain}.}
    \label{tab:abl_train}
\end{table}

%% file: tab/pseudo_label_plus_coco.tex
\setlength{\tabcolsep}{0.006\linewidth}{
\begin{table*}[t]
\scriptsize
\RawFloats
\centering
\makebox[0pt][c]{\parbox{1\textwidth}{%
    \begin{minipage}[b]{0.49\hsize}\centering
    \begin{tabular}{ccc} 
        \toprule
        PL & Recall (IoU=0.5) & Recall (IoU=0.7) \\ \midrule 
        Initial  & 0.5422 & 0.4671 \\
        Round 1  & 0.8492 & 0.7192  \\
        Round 2  & 0.8789 & 0.7422  \\
        Round 3 & 0.8795 & 0.7421  \\
        \bottomrule
    \end{tabular}
    \vspace{-2.5mm}
    \caption{\textbf{Quality of the pseudo-label} on the KITTI validation set. PL notes the Pseudo Label.
    }
    \label{tab:quality}
       
    \end{minipage}
    \hfil
    \begin{minipage}[b]{0.46\hsize}\centering
\begin{tabular}{llccc}
    \toprule
      \multirow{2}{*}{Method}& \multirow{2}{*}{2D Anno.}& \multicolumn{3}{c}{\apthreeD@IoU=0.7}  \\
    \cmidrule(lr){3-5}  & & Easy & Mod. & Hard \\
    \midrule
    FGR\cite{wei2021fgr} & KITTI & 86.68 & 73.55 & 67.91 \\
    VG-W3D (Ours) & KITTI & 91.32 & 78.89 & 74.70 \\
    \midrule
    FGR\cite{wei2021fgr}
    & COCO & 88.08 & 73.87 & 68.69   \\
    VG-W3D (Ours) & COCO & 89.12 & 78.21 & 74.21\\
    \bottomrule
    \end{tabular}
        \vspace{-2.5mm}
        \caption{\textbf{Integration with the off-the-shelf 2D object detector}, which is pre-trained on a different dataset.}
    \label{tab:coco}

    \end{minipage}
    \hfil
}}
\end{table*}
}

%% file: sec/4_Conclusion.tex
\section{Conclusions}
In this paper,  we introduce a multi-level visual guidance approach for weakly-supervised 3D object detection by only using 2D annotations.  
Leveraging rich information from the image domain, our method employs visual data to guide the 3D learning process from feature-, output-, and training-level perspectives. 
At the feature level, our guidance ensures that point cloud features predict objectness distributions similar to those of the image detector. Output-level guidance enforces the LiDAR model to generate proposals located at reasonable positions. 
Moreover, training-level guidance generates high-quality 3D labels consistent with 2D bounding boxes, facilitating the retraining process. 
Extensive experiments conducted on the KITTI dataset demonstrate that our method outperforms state-of-the-art approaches, achieving competitive performance without needing any 3D annotations.